\documentclass[10pt,twocolumn,letterpaper]{article}

\usepackage{cvpr}
\usepackage{times}
\usepackage{epsfig}
\usepackage{graphicx}
\usepackage{amsmath}
\usepackage{amssymb}
\usepackage{booktabs}
\usepackage{multirow}
\usepackage{bm}
\graphicspath{{fig}}
\usepackage{subcaption}
\graphicspath {{fig/}}
\usepackage{verbatim} 

\usepackage{color}

\DeclareMathOperator*{\argmax}{arg\,max} 

\usepackage[breaklinks=true,bookmarks=false]{hyperref}

\cvprfinalcopy 

\def\cvprPaperID{2790} 

\ifcvprfinal\fi
\begin{document}

\title{Defense against Adversarial Attacks Using \\High-Level Representation Guided Denoiser}

\author{Fangzhou Liao\thanks{Equal contribution.}, Ming Liang\footnotemark[1], Yinpeng Dong, Tianyu Pang, Xiaolin Hu\thanks{Corresponding author.}, Jun Zhu\\
 Department of Computer Science and Technology, Tsinghua Lab of Brain and Intelligence,\\
 Beijing National Research Center for Information Science and Technology, BNRist Lab\\
 Tsinghua University, 100084 China\\
\small{\{liaofangzhou, liangming.tsinghua\}@gmail.com, \{dyp17, pty17\}@mails.tsinghua.edu.cn, \{xlhu, dcszj\}@tsinghua.edu.cn}
}

\maketitle

\begin{abstract}
Neural networks are vulnerable to adversarial examples, which poses a threat to their application in security sensitive systems. We propose high-level representation guided denoiser (HGD) as a defense for image classification. Standard denoiser suffers from the error amplification effect, in which small residual adversarial noise is progressively amplified and leads to wrong classifications. HGD overcomes this problem by using a loss function defined as the difference between the target model's outputs activated by the clean image and denoised image. Compared with ensemble adversarial training which is the state-of-the-art defending method on large images, HGD has three advantages. First, with HGD as a defense, the target model is more robust to either white-box or black-box adversarial attacks. Second, HGD can be trained on a small subset of the images and generalizes well to other images and unseen classes. Third, HGD can be transferred to defend models other than the one guiding it. In NIPS competition on defense against adversarial attacks, our HGD solution won the first place and outperformed other models by a large margin. \footnote{Code: \url{https://github.com/lfz/Guided-Denoise}.}
\end{abstract}

\section{Introduction}
As many other machine learning models \cite{biggio2013evasion}, neural networks are known to be vulnerable to adversarial examples \cite{szegedy2013intriguing,goodfellow2014explaining}. Adversarial examples are maliciously designed inputs to attack a target model. They have small perturbations on original inputs but can mislead the target model. Adversarial examples can be transferred across different models \cite{szegedy2013intriguing,papernot2016transferability}. This transferability enables black-box adversarial attacks without knowing the weights and structures of the target model. Black-box attacks have been shown to be feasible in real-world scenarios \cite{papernot2017practical}, which poses a potential threat to security-sensitive deep learning applications, such as identity authentication and autonomous driving. It is thus important to find effective defenses against adversarial attacks.

\begin{figure}[t!]
\includegraphics[width=\columnwidth]{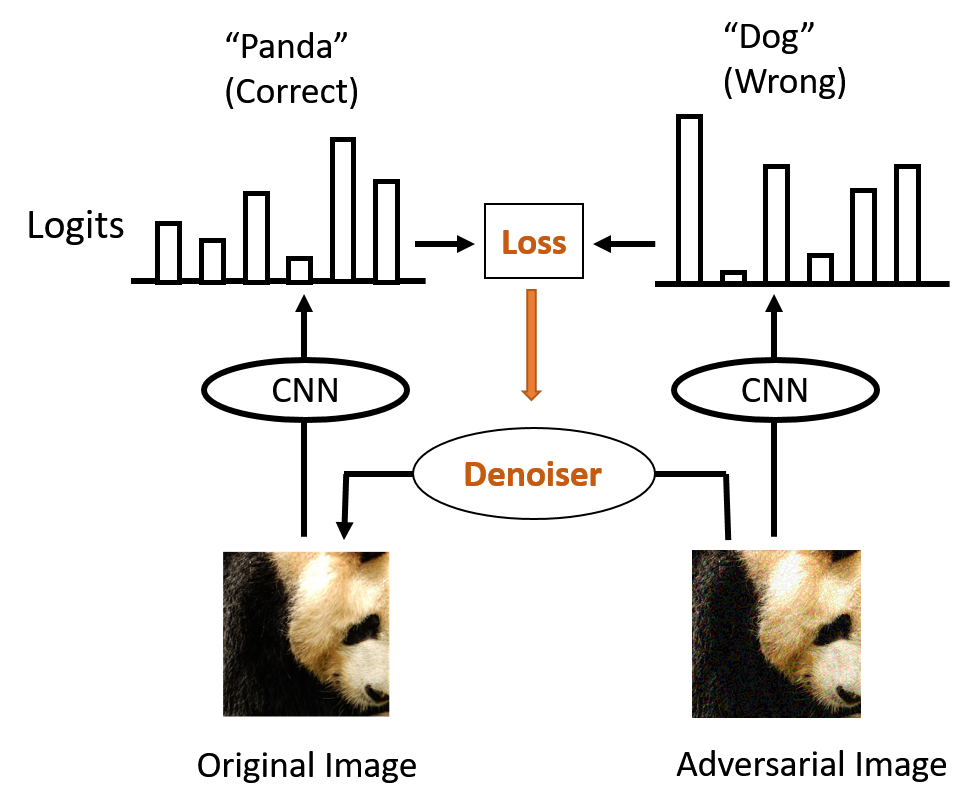}
\caption{The idea of high-level representation guided denoiser. The difference between the original image and adversarial image is tiny, but the difference is amplified in high-level representation (logits for example) of a CNN. We use the distance over high-level representations to guide the training of an image denoiser to suppress the influence of adversarial perturbation. }
\label{fig_abs}
\end{figure}

Since adversarial examples are constructed by adding noises to original images, a natural idea is to denoise adversarial examples before sending them to the target model (Figure \ref{fig_abs}). We explored two models for denoising adversarial examples, and found that the noise level could indeed be reduced. These results demonstrate the feasibility of the denoising idea. However, none of the models can remove all adversarial perturbations, and small residual perturbation is amplified to a large magnitude in top layers of the target model (called ``error amplification effect"), which leads to a wrong prediction. To solve this problem, instead of using a pixel-level reconstruction loss function as standard denoisers, we set the loss function as the difference between top-level outputs of the target model induced by original and adversarial examples (Figure \ref{fig_abs}). We name the denoiser trained by this loss function ``high-level representation guided denoiser" (HGD). 



Compared with ensemble adversarial training \cite{tramer2017ensemble} which is the current state-of-the-art method, the proposed method has the following advantages. First, it achieves much higher accuracy when defending both white-box and black-box attacks. Second, HGD requires much less training data and training time, and well generalizes to other images and unseen classes. Third, HGD can be transferred across different target models. We further validated the performance of HGD in the NIPS adversarial defense competition. Our HGD approach won the first place by a large margin, and had faster inference speed than other top-ranked methods.


\section{Background and Related Work}\label{sec:background}
In this section, we first specify some of the notations used in this paper.
Let $x$ denote the clean image from a given dataset, and $y$ denote the class. The ground truth label is denoted by $y_{\text{true}}$.
A neural network $f:x\rightarrow y$ is called the target model. Given an input $x$, its feature vector at layer $l$ is $f_l(x)$, and its predicted probability of class $y$ is $p(y|x)$.
$y_{x} = \argmax_yp(y|x)$ is the predicted class of $x$. $J(x, y)$ denotes the loss function of the classifier given the input $x$ and its target class $y$. For image classification, $J(x, y)$ is often chosen to be the cross-entropy loss.
We use $x^*$ to denote the adversarial example generated from $x$. 
$\epsilon$ is the magnitude of adversarial perturbation, measured by a certain distance metric.

\subsection{Existing methods for adversarial attacks}
Adversarial examples \cite{szegedy2013intriguing} are maliciously designed inputs which have a small difference from clean images but cause the classifier to give wrong classifications. That is, for $x^*$ with a sufficiently small perturbation magnitude $\epsilon$ , $y_{x^*} \neq y_{x}$. We use $L_\infty$ to measure $\epsilon$ in this study.

Szegedy et al.~\cite{szegedy2013intriguing} use a box-constrained L-BFGS algorithm to generate targeted adversarial examples, which bias the predictions to a specified class $y_{target}$. More specifically, they minimize the weighted sum of $\epsilon$ and $J(x^*, y_{target})$ while constraining the elements of $x^*$ to be normal pixel value. 

Goodfellow et al. \cite{goodfellow2014explaining} suggest that adversarial examples can be caused by the cumulative effects of high dimensional model weights. They propose a simple adversarial attack algorithm, called Fast Gradient Sign Method (FGSM):
\begin{equation}\label{equ:fgsm}
x^* = x + \epsilon \cdot \text{sign}(\nabla_{x}J(x, y)).
\end{equation}
FGSM only computes the gradients for once, and thus is much more efficient than L-BFGS. In early practices, FGSM uses the true label $y = y_{\text{true}}$ to compute the gradients. This approach is suggested to have the label leaking \cite{kurakin2016adversarial} effect, in that the generated adversarial example contains the label information. A better alternative is to replace $y_\text{true}$ with the model predicted class $y_{x}$. FGSM is untargeted and aims to increase the overall loss. Targeted FGSM can be obtained by modifying FGSM to maximize the predicted probability of a specified class $y_{\text{target}}$:
\begin{equation}\label{equ:tfgsm}
x^* = x - \epsilon \cdot \text{sign}(\nabla_{x}J(x, y_\text{target}))
\end{equation}
$y_\text{target}$ can be chosen as the least likely class predicted by the model or a random class. 
Kurakin et al.\cite{kurakin2016adversarial} propose an iterative FGSM (IFGSM) attack by repeating FGSM for $n$ steps (IFGSMn). IFSGM usually results in higher classification error than FGSM. 


The model used to generate adversarial attacks is called the attacking model, which can be a single model or an ensemble of models \cite{tramer2017ensemble}. When the attacking model is the target model itself or contains the target model, the resulting attacks are white-box. An intriguing property of adversarial examples is that they can be transferred across different models \cite{szegedy2013intriguing,goodfellow2014explaining}. This property enables black-box attacks. Practical black-box attacks have been demonstrated in some real-world scenarios \cite{papernot2017practical,papernot2016transferability}. As white-box attacks are less likely to happen in practical systems, defenses against black-box attacks are more desirable.


\subsection{Existing methods for defenses}
Adversarial training \cite{goodfellow2014explaining,kurakin2016adversarial,tramer2017ensemble} is one of the most extensively investigated defenses against adversarial attacks. It aims to train a robust model from scratch on a training set augmented with adversarially perturbed data \cite{goodfellow2014explaining,kurakin2016adversarial,tramer2017ensemble}. 
Adversarial training improves the classification accuracy of the target model on adversarial examples \cite{szegedy2013intriguing,goodfellow2014explaining,kurakin2016adversarial,tramer2017ensemble}. On some small image datasets it even improves the accuracy of clean images \cite{szegedy2013intriguing,goodfellow2014explaining}, although this effect is not found on ImageNet \cite{deng2009imagenet} dataset. However, adversarial training is more time consuming than training on clean images only, because online adversarial example generation needs extra computation, and it takes more epochs to fit adversarial examples \cite{tramer2017ensemble}.
These limitations hinder the usage of harder attacks in adversarial training, and practical adversarial training on the ImageNet dataset only adopts FGSM.

Preprocessing based methods process the inputs with certain transformations to remove the adversarial noise, and then send these inputs to the target model. Gu and Rigazio \cite{gu2014towards} first propose the use of denoising auto-encoders as a defense. Osadchy et al. \cite{osadchy2016no} apply a set of filters to remove the adversarial noise, such as the median filter, averaging filter and Gaussian low-pass filter. Graese et al. \cite{graese2016assessing} assess the defending performance of a set of preprocessing transformations on MNIST digits \cite{lecun1998gradient}, including the perturbations introduced by image acquisition process, fusion of crops and binarization. Das et al. \cite{das2017keeping} preprocess images with JPEG compression to reduce the effect of adversarial noises. Meng and Chen \cite{meng2017magnet} propose a two-step defense model, which detects the adversarial input and then reform it based on the difference between the manifolds of clean and adversarial examples. Our approach distinguishes from these methods by using the reconstruction error of high-level features to guide the learning of denoisers. Moreover, these methods are usually evaluated on small images. As we will show in experiments section, some method effective on small images may not transfer well to large images.

Another family of adversarial defenses is based on the so-called gradient masking effect \cite{papernot2017practical,papernot2016towards,tramer2017ensemble}. These defenses apply some regularizers or smooth labels to make the model output less sensitive to the perturbation on input. Gu and Rigazio \cite{gu2014towards} propose the deep contrastive network, which uses a layer-wise contrastive penalty term to achieve output invariance to input perturbation. Papernot et al. \cite{papernot2016distillation} adapts knowledge distillation \cite{hinton2015distilling} to adversarial defense, and uses the output of another model as soft labels to train the target model. Nayebi and Surya \cite{nayebi2017biologically} use saturating networks for robustness to adversarial noises. The loss function is designed to encourage the activations to be in their saturating regime. The basic problem with these gradient masking approaches is that they fail to solve the vulnerability of the models to adversarial attacks, but just make the construction of white-box adversarial examples more difficult. These defenses still suffer from black-box attacks \cite{papernot2017practical,tramer2017ensemble} generated on other models.

\section{Methods}

\subsection{Pixel guided denoiser}\label{sec:model}

In this section, we introduce a set of denoising networks and their motivations. These denoisers are designed in the context of image classification on ImageNet \cite{deng2009imagenet} dataset. They are used in conjunction with a pretrained classifier $f$ (By default Inception V3 \cite{szegedy2016rethinking} in this study). Let $x$ denote the clean image. The denoising function is denoted as $D:x^* \rightarrow \hat{x}$, where $x^*$ and $\hat{x}$ denote the adversarial image and denoised image, respectively. The loss function is:
\begin{equation}
L = ||x-\hat{x}||,
\end{equation}
where $||\cdot||$ stands for the $L_1$ norm, the following equations also use this notation.
Since the loss function is defined at the level of image pixels, we name this kind of denoiser pixel guided denoiser (PGD).

\begin{figure}
\centering
\includegraphics[width=\columnwidth]{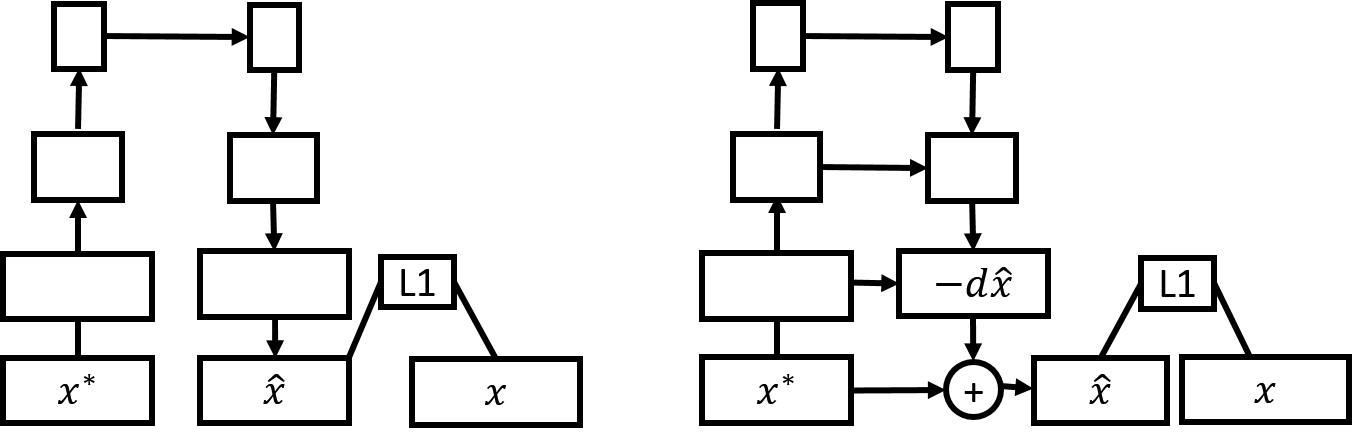}
\caption{Diagrams of DAE (left) and DUNET (right).}
\label{fig_strct}
\end{figure}

\begin{figure*}
\includegraphics[width=0.9\linewidth]{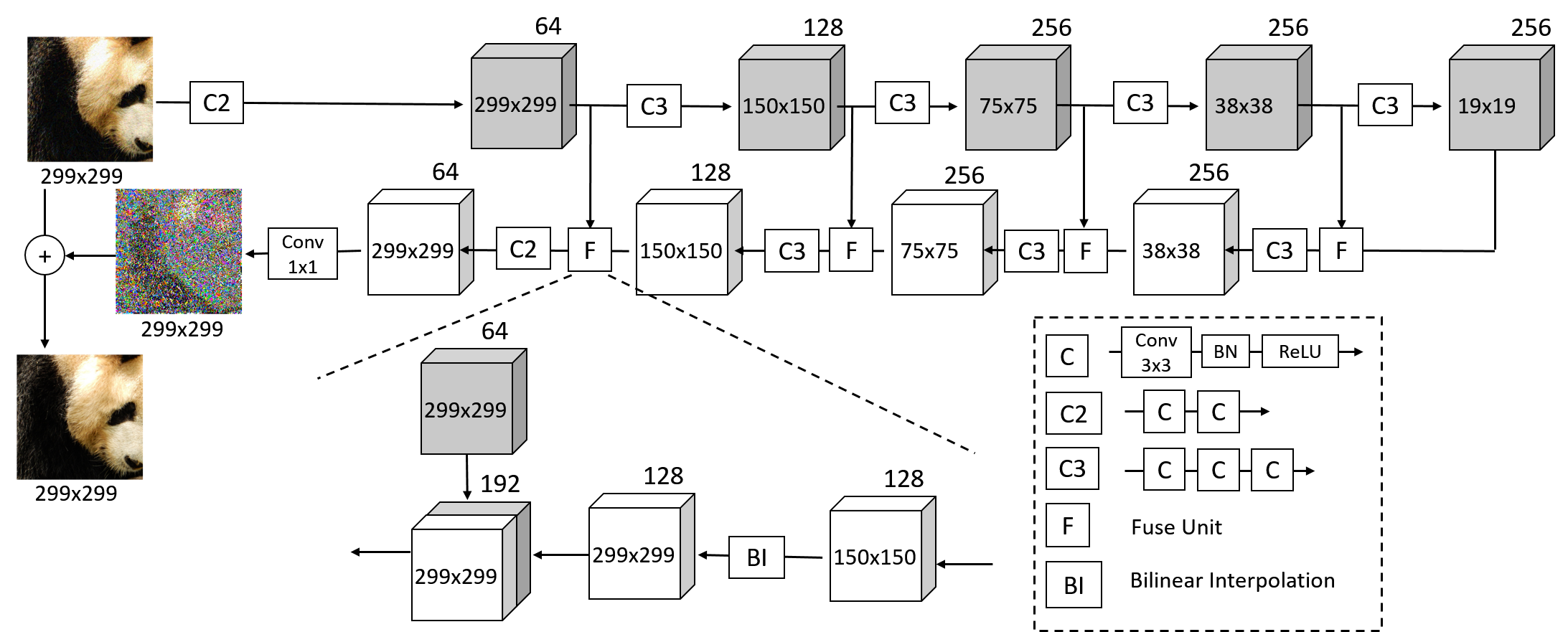}
\caption{
The detail of DUNET. The numbers inside each cube stand for width $\times$ height, and the number outside the cube stands for the number of channels. In all the C3 of the feedforward path, the stride of the first C is $2\times2$.}
\label{fig_daunet}
\end{figure*}

\subsubsection{Denoising U-net}
Denoising autoencoder (DAE) \cite{vincent2008extracting} is a popular denoising model. In a previous work \cite{gu2014towards}, DAE in the form of a multi-layer perceptron was used to defend target models against adversarial attacks. However, the experiments were conducted on the relatively simple MNIST \cite{lecun1998gradient} dataset.
To better represent the high-resolution images in the ImageNet dataset, we use a convolutional version of DAE for the experiments (see Figure \ref{fig_strct} left). 

DAE has a bottleneck structure between the encoder and decoder. This bottleneck may constrain the transmission of fine-scale information necessary for reconstructing high-resolution images. To overcome this problem, we modify DAE with the U-net \cite{ronneberger2015u} structure and propose the denoising U-net (DUNET, see Figure \ref{fig_strct} right). DUNET is different from DAE in two aspects. First, similar to the Ladder network \cite{rasmus2015semi}, DUNET adds lateral connections from encoder layers to their corresponding decoder layers in the same scale. Second, the learning objective of DUNET is the adversarial noise ($d\hat{x}$ in Figure \ref{fig_strct}), instead of reconstructing the whole image as in DAE. This residual learning \cite{zhang2017beyond} is implemented by the shortcut from input to output to additively combine them. The clean image can be readily obtained by subtracting the noise (adding -$d\hat{x}$) from the corrupted input.

\subsubsection{Network architecture}
We use DUNET as an example to illustrate the architecture (Figure \ref{fig_daunet}). DAE can be obtained simply by removing the lateral connections from DUNET.
$C$ is defined as a stack of layer sequences, and each sequence contains a $3\times3$ convolution, a batch normalization layer \cite{ioffe2015batch} and a rectified linear unit. $Ck$ is defined as $k$ consecutive $C$. The network is composed of a feedforward path and a feedback path. The feedforward path is composed of five blocks, corresponding to one $C2$ and four $C3$, respectively. The first convolution of each $C3$ has $2\times 2$ stride, while the stride of all other layers is $1\times 1$. The feedforward path receives the image as input, and generates a set of feature maps of increasingly lower resolutions (see the top pathway of Figure \ref{fig_daunet}).

The feedback path is composed of four blocks and a $1\times 1$ convolution. Each block receives a feedback input from the feedback path and a lateral input from the feedforward path. It first upsamples the feedback input to the same size as the lateral input using bilinear interpolation, and then processes the concatenated feedback and lateral inputs with a $Ck$. From top to bottom, three $C3$ and one $C2$ are used. Along the feedback path, the resolution of feature maps is increasingly higher. The output of the last block is transformed to the negative noise $-d\hat{x}$ by a $1\times 1$ convolution (See the bottom pathway of Figure \ref{fig_daunet}). The final output is the sum of the negative noise and the input image:
\begin{equation}
\hat{x}=x^*-d\hat{x}.
\label{eq:xhat}
\end{equation}


\begin{figure*}
\centering
\begin{subfigure}[t]{0.25\linewidth}
\includegraphics[width=\columnwidth]{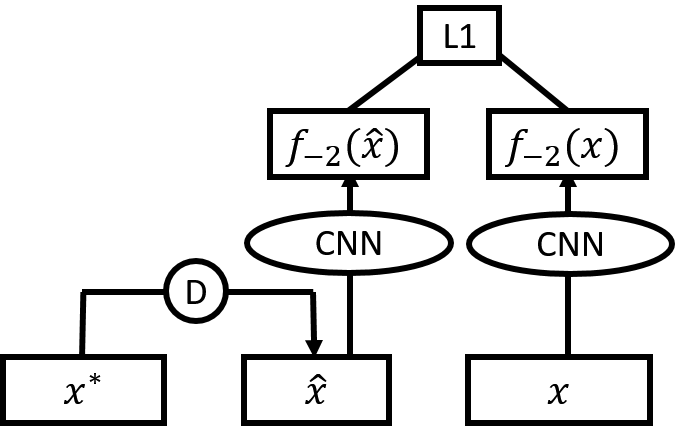}
\subcaption{FGD}
\end{subfigure}
\hspace{1cm}
\begin{subfigure}[t]{0.25\linewidth}
\includegraphics[width=\columnwidth]{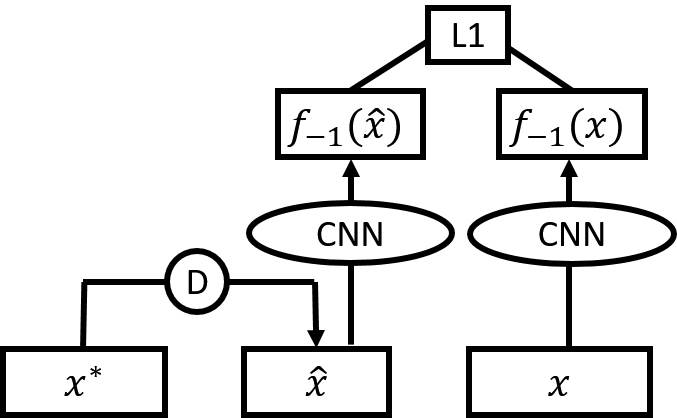}
\subcaption{LGD}
\end{subfigure}
\hspace{1cm}
\begin{subfigure}[t]{0.25\linewidth}
\includegraphics[width=\columnwidth]{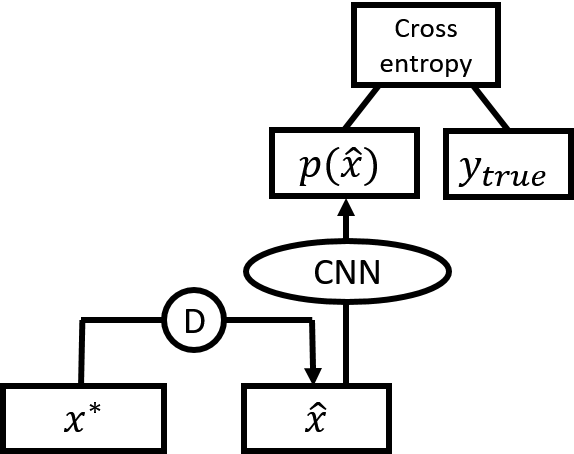}
\subcaption{CGD}
\end{subfigure}
\caption{Three different training methods for HGD. The square boxes stand for data blobs, the circles and ovals stand for networks. D stands for denoiser. CNN is the model to be defended. The parameters of the CNN are shared and fixed.}\label{fig:denoiser}
\end{figure*}

\subsection{High-level representation guided denoiser}
A potential problem with PGD is the amplification effect of adversarial noise. Adversarial examples have negligible differences from the clean images. However, this small perturbation is progressively amplified by deep neural networks and yields a wrong prediction. Even if the denoiser can significantly suppress the pixel-level noise, the remaining noise may still distort the high-level responses of the target model. Refer to Section \ref{sec:pgd} for details.

To overcome this problem, we replace the pixel-level loss function with the reconstruction loss of the target model's outputs. More specifically, given a target neural network, we extract its representations at $l$-th layer activated by $x$ and $\hat{x}$, and define the loss function as the $L_1$ norm of their difference:
\begin{equation}
L=||f_l(\hat{x}) - f_l(x)||.
\end{equation}
The corresponding model is called HGD, in that the supervised signal comes from certain high-level layers of the target model and carries guidance information related to image classification. HGD uses the same U-net structure as DUNET. They only differ in their loss functions.

We propose two HGDs with different choices of $l$. For the first HGD, we define $l=-2$ as the index of the topmost convolutional layer. The activations of this layer are fed to the linear classification layer after global average pooling, so it is more related to the classification objective than lower convolutional layers. This denoiser is called feature guided denoiser (FGD) (see Figure \ref{fig:denoiser}a). The loss function used by FGD is also known as perceptual loss or feature matching loss\cite{ridgeway2015learning,johnson2016perceptual,dosovitskiy2016generating}. For the second HGD, we define $l=-1$ as the index of the layer before the final softmax function, i.e., the logits. This denoiser is called logits guided denoiser (LGD). In this case, the loss function is the difference between the two logits activated by $\hat{x}$ and $x$ (see Figure \ref{fig:denoiser}b).
We consider both FGD and LGD for the following reason. The convolutional feature maps provide richer supervised information, while the logits directly represent the classification results.

All PGD and these HGDs are unsupervised models, in that the ground truth labels are not needed in their training process. An alternative is to use the classification loss of the target model as the denoising loss function, which is supervised learning as ground truth labels are needed. This model is called class label guided denoiser (CGD) (see Figure\ref{fig:denoiser}c).

\section{Experimental settings}\label{sec:exp}

Throughout experiments, the pretrained Inception v3 (IncV3) \cite{szegedy2016rethinking} is assumed to be the target model that attacks attempt to fool and our denoisers attempt to defend. Therefore this model is used for training the three HGDs illustrated in Figure \ref{fig:denoiser}. However, it will be seen that the HGDs trained with this target model can also defend other models (see Section \ref{subsec:transfer}). All our experiments are conducted on images from the ImageNet dataset. Although many defense methods have been proposed, they are mostly evaluated on small images and only adversarial training is thoroughly evaluated on ImageNet. We compare our model with ensemble adversarial training, which is the state-of-the-art defense method of defending large images.

\subsection{Dataset}
For both training and testing of the proposed method, adversarial images are needed. To prepare the training set, we first extract 30K images from the ImageNet training set (30 images per class). Then we use a bunch of adversarial attacking methods to distort these images and form a training set of adversarial images. Different attacking methods including FGSM and IFGSM are applied to the following models:  
  Pre-trained IncV3, InceptionResnet v2 (IncResV2) \cite{szegedy2017inception}, ResNet50 v2 (Res) \cite{he2016identity} individually or in combinations (the same model ensemble as the work of Tramer et al. \cite{tramer2017ensemble} ). For each training sample, the perturbation level $\epsilon$ is uniformly sampled from integers in $[1,16]$. See Table \ref{tab_attacklist} for details. As a consequence, we gather 210K images in the training set (TrainSet). 
  
To prepare the validation set, we first extract 10K images from the ImageNet training set (10 images per class), then apply the same method as described above. Therefore the size of the validation set (ValSet) is 70K.   

Two different test sets are constructed, one for white-box attacks (WhiteTestSet)\footnote{The white-box attacks defined in this paper should be called oblivious attacks according to Carlini and Wagner's definition \cite{carlini2017adversarial}} and the other for black-box attacks (BlackTestSet). They are obtained from the same clean 10K images from the ImageNet validation set (10 images per class) but using different attacking methods. The WhiteTestSet uses two attacks targeting at IncV3, which is also used for generating training images, and the BlackTestSet uses two attacks based on a holdout model pre-trained Inception V4 (IncV4) \cite{szegedy2017inception}, which is not used for generating training images. Every attacking method is conducted on two perturbation levels $\epsilon \in \{4, 16\}$. So both WhiteTestSet and BlackTestSet have 40k images (see Table \ref{tab_attacklist} for details).

\begin{table}[]
\centering
\caption{Adversarial images generated by different models for training and testing.}
\label{tab_attacklist}
\small
\begin{tabular}{llll}
\hline
                 & \begin{tabular}[c]{@{}l@{}}Attacking \\ method \end{tabular} & Attacked model & $\epsilon$                                                                     \\ \hline
\multirow{8}{*}{\begin{tabular}[c]{@{}l@{}}TrainSet \\ and ValSet \end{tabular} } & FGSM             & IncV3         & \multirow{8}{*}{[1,16]}                                                                                       \\
                           & FGSM             & IncResV2                                                                                             \\
                           & FGSM             & Res                                                                                                  \\
                           & FGSM             & IncV3/IncResV2/Res                                                                                \\
                           & IFGSM2            & IncV3/IncResV2/Res                                                                               \\
                           & IFGSM4            & IncV3/IncResV2/Res                                                                               \\ 
                           & IFGSM8            & IncV3/IncResV2/Res                                                                               \\ \hline
\multirow{2}{*}{WhiteTestSet} & FGSM             & IncV3          & \multirow{2}{*}{\{4,16\}}                                                                                      \\
                           & IFGSM4            & IncV3/IncResV2/Res                                                                               \\ \hline
\multirow{2}{*}{BlackTestSet}  & FGSM             & IncV4        & \multirow{2}{*}{\{4,16\}}              \\
                           & IFGSM4            & IncV4                                                                                                \\ \hline
\end{tabular}
\end{table}

\subsection{Implementation details}
The denoisers are optimized using Adam \cite{kingma2014adam}. The learning rate is initially set to 0.001, and decay to 0.0001 when the training loss converges. The model is trained on six GPUs and the batch size is 60. The number of training epochs ranges from 20 to 30, depending on the convergence speed of the model. The model with the lowest validation loss is used for testing.

\begin{table}[]
\centering
\caption{Denosing loss and classification accuracy of different PGD methods on the test sets. Denosing loss is the $L_1$ distance between the input image and the denoised image. NA means no defense. Clean stands for original images.}
\label{tab_px}
\small
\begin{tabular}{cccccc}
\hline
\multirow{2}{*}{Defense} & \multirow{2}{*}{Clean} & \multicolumn{2}{c}{WhiteTestSet}  & \multicolumn{2}{c}{BlackTestSet} \\
                         &                        & $\epsilon=4$ & $\epsilon=16$ & $\epsilon=4$ & $\epsilon=16$ \\ \hline
  NA & \textbf{0.0000}\ &     0.0177 &      0.0437 &      0.0176 &       0.0451 \\
    DAE & 0.0360 &     0.0359 &      0.0360 &      0.0360 &       0.0369 \\
 DUNET & 0.0150 &     \textbf{0.0140} &      \textbf{0.0164} &      \textbf{0.0140} &       \textbf{0.0181} \\
\hline
\hline
  NA &  \textbf{76.7\%} &      14.5\% &       14.4\% &       61.2\% &        41.0\% \\
    DAE &  58.3\% &      \textbf{51.4\%} &       \textbf{36.7}\% &       55.9\% &        48.8\% \\
 DUNET &  75.3\% &      20.0\% &       13.8\% &       \textbf{67.5\%} &        \textbf{55.7\%} \\
 \hline
\end{tabular}
\end{table}

\begin{figure}
\centering
\includegraphics[width=0.8\columnwidth]{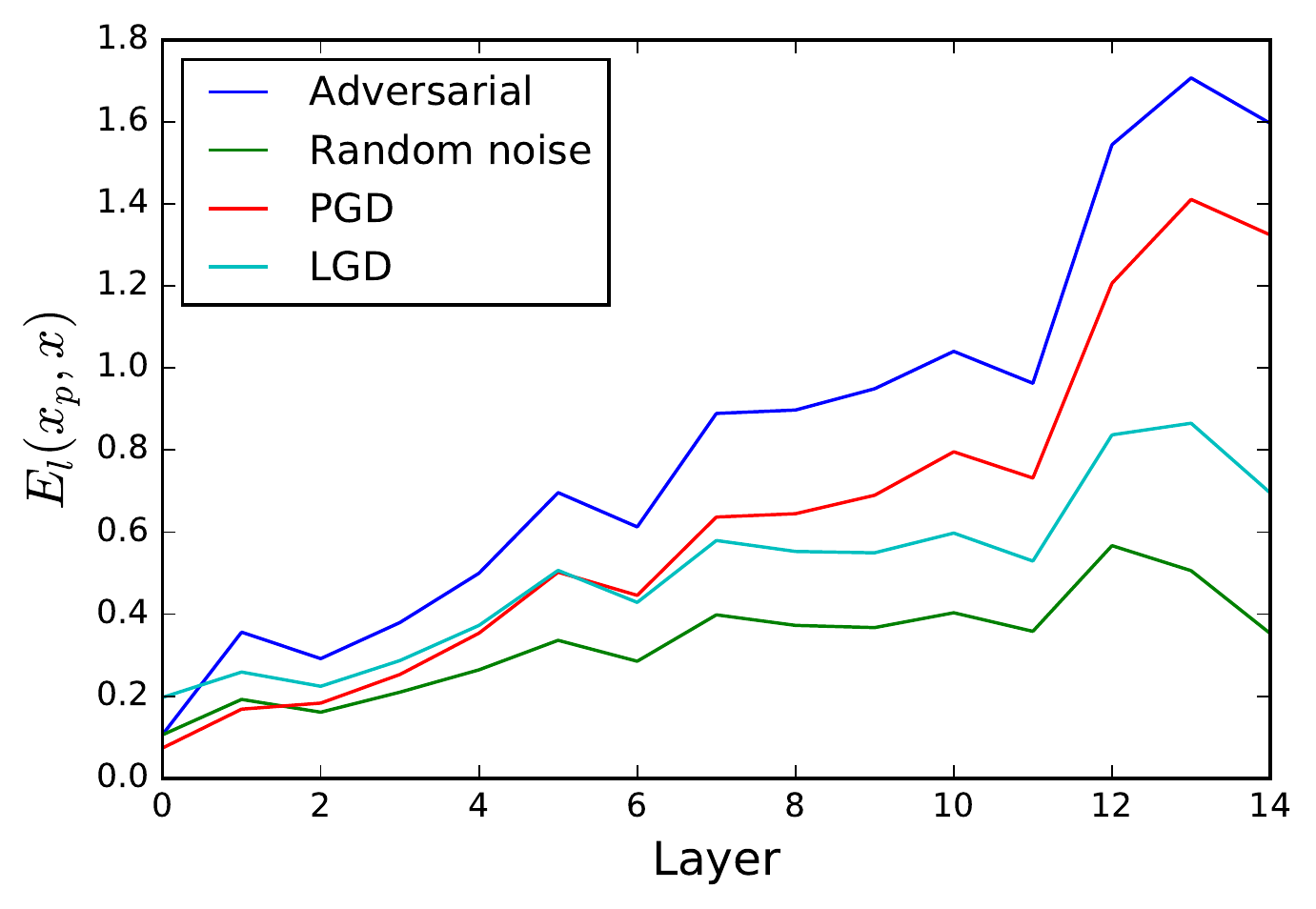}
\caption{Layerwise perturbation levels of the target model.  Adversarial, Random noise, PGD and LGD correspond to the $E_l$ for adverarial images, Gaussian noise perturbed images, PGD denoised images, and LGD denoised images, respectively.}
\label{fig_amp}
\end{figure}

\section{Results}\label{sec:result}

\subsection{PGD and the error amplification effect} \label{sec:pgd}
The results of DAE and DUNET on the test sets are shown in Table \ref{tab_px}\footnote{For detailed results of each attack in Table 2-5, please refer to the supplementary material.}. The original IncV3 without any defense is used as a baseline, denoted as NA. For all types of attacks, DUNET has much lower denoising loss than DAE and NA, which demonstrates the structural advantage of DUNET. DAE does not perform well on encoding the high-resolution images, as its accuracy on clean images significantly drops. DUNET slightly decreases the accuracy of clean images, but significantly improves the robustness of the target model to black-box attacks. In what follows, DUNET is used as the default PGD method.

A notable result is that the denoising loss and classification accuracy of PGD are not so consistent. For white-box attacks, DUNET has much lower denoising loss than DAE, but its classification accuracy is significantly worse. To investigate this inconsistency, we analyze the layer-wise perturbations of the target model activated by PGD denoised images. Let $x_p$ denote a perturbed image. The perturbation level at layer $l$ is computed as:
\begin{align}
E_l(x_p,x) &= || f_l(x_p) - f_l(x)|| / ||f_l(x)||.
\end{align}

The $E_l$ for PGD denoised images, adversarial images, and Gaussian noise perturbed images are shown in Figure \ref{fig_amp}. The latter two are used as baselines. The curves are the averaged results on 30 randomly picked adversarial images generated by the ensemble attack "IFGSM4 x IncV3/IncResV2/Res ($\epsilon=16$)". For convenience, these $E_l$ are abbreviated as PGD perturbation, adversarial perturbation, and random perturbation. Although the pixel-level PGD perturbation significantly suppressed, the remaining perturbation is progressively amplified along the layer hierarchy. At the top layer, PGD perturbation is much higher than random perturbation and close to adversarial perturbation. Because the classification result is closely related to the top-level features, this large perturbation well explains the inconsistency between the denoising performance and classification accuracy of PGD.

\begin{table}[]
\centering
\caption{The classification accuracy on test sets obtained by different defenses. NA means no defense.}
\label{tab_gd}
\small
\begin{tabular}{cccccc}
\hline
\multirow{2}{*}{Defense} & \multirow{2}{*}{Clean} & \multicolumn{2}{c}{WhiteTestSet}  & \multicolumn{2}{c}{BlackTestSet} \\
                         &                        & $\epsilon=4$ & $\epsilon=16$ & $\epsilon=4$ & $\epsilon=16$ \\ \hline
  NA &  76.7\% &      14.5\% &       14.4\% &       61.2\% &        41.0\% \\
  
  PGD &  75.3\% &      20.0\% &       13.8\% &       67.5\% &        55.7\% \\
   ensV3 \cite{tramer2017ensemble} &  \textbf{76.9\%} &      69.8\% &       58.0\% &       72.4\% &        62.0\% \\
 \hline
    
  FGD &  76.1\% &      73.7\% &       67.4\% &       74.3\% &        71.8\% \\
 LGD &  76.2\% &      75.2\% &       69.2\% &       \textbf{75.1\%} &        \textbf{72.2}\% \\
 CGD &  74.9\% &      \textbf{75.8\%} &               \textbf{73.2\%} &       74.5\% &        71.1\% \\
 \hline
 \end{tabular}
\end{table}

\subsection{Evaluation results of HGD}
Compared to PGD, LGD strongly suppress the error amplification effect (Figure \ref{fig_amp}). LGD perturbation at the final layer is much lower than PGD and adversarial perturbations and close to random perturbation. 

HGD is more robust to white-box and black-box adversarial attacks than PGD and ensV3 (Table \ref{tab_gd}). All three HGD methods significantly outperform PGD and ensV3 for all types of attacks. The accuracy of clean images only slightly decreases (by $0.5\%$ for LGD). The difference between these HGD methods is insignificant. In later sections, LGD is chosen as our default HGD method for it achieves a good balance between accuracy on clean and adversarial images. When facing powerful ensemble black-box attacks, LGD also significantly outperforms ensV3 (see supplementary material).

Compared to adversarial training, HGD only uses a small fraction of training images and is efficient to train. Only 30K clean images are used to construct our training set, while all 1.2M clean images of the ImageNet dataset are used for training ensV3. HGD is trained for less than 30 epochs on 210K adversarial images, while ensV3 is trained for about 200 epochs on 1.2M images \cite{tramer2017ensemble}.

To summary, with less training data and time, HGD significantly outperforms adversarial training on defense against adversarial attacks. These results suggest that learning to denoise only is much easier than learning the coupled task of classification and defense.

\subsection{Transferability of HGD}\label{subsec:transfer}
The learning objective of HGD is to remove the high-level influence of adversarial noises. In other words, HGD works by producing anti-adversarial perturbations on input images. From this point of view, we expect that HGD can be transfered to defend other models and images. 

To evaluate the transferability of HGD over different models, we use the IncV3 guided LGD to defend Resnet \cite{he2016deep}. As expected, this LGD significantly improves the robustness of Resnet to all attacks. Furthermore, it achieves very close defending performance as the Resnet guided LGD. (Table \ref{tab_trans2})

To evaluate the transferability of HGD over different classes, we build another dataset. Its key difference from the original dataset is that there are only 750 classes in the TrainSet, and the other 250 classes are put in ValSet and TestSets. The number of original images in each class in all datasets are changed to 40 to keep the size of dataset unchanged. It is found that although the 250 classes in the test set are never trained, the LGD still learns to defend against the attacks targeting at them (Table \ref{tab_trans}).

\begin{table}
\caption{The transferability of HGD to different model. Resnet is used as the target model.}
\label{tab_trans2}
\small
\begin{tabular}{cccccc}
\hline
\multirow{2}{*}{\begin{tabular}[c]{@{}c@{}} Denoiser for\\Resnet \end{tabular}} & \multirow{2}{*}{Clean} & \multicolumn{2}{c}{WhiteTestSet}  & \multicolumn{2}{c}{BlackTestSet} \\
                         &                        & $\epsilon=4$ & $\epsilon=16$ & $\epsilon=4$ & $\epsilon=16$ \\ \hline
    NA &  \textbf{78.5}\% &      63.3\% &       38.4\% &       67.8\% &        48.6\% \\
   \begin{tabular}[c]{@{}c@{}} IncV3 guided \\LGD \end{tabular} &  77.4\% &      75.8\% &       71.7\% &       76.1\% &        72.7\% \\
\begin{tabular}[c]{@{}c@{}} Resnet guided \\LGD \end{tabular} &  78.4\% &      \textbf{76.1\%} &      \textbf{72.9\%} &       \textbf{76.5\%} &        \textbf{74.6\%} \\
  \hline
\end{tabular}
\end{table}

\begin{table}
\caption{The transferability of HGD to different classes. The 1000 ImageNet classes are separated in training and test test.}
\label{tab_trans}
\small
\begin{tabular}{cccccc}
\hline
\multirow{2}{*}{Defense} & \multirow{2}{*}{Clean} & \multicolumn{2}{c}{WhiteTestSet}  & \multicolumn{2}{c}{BlackTestSet} \\
                         &                        & $\epsilon=4$ & $\epsilon=16$ & $\epsilon=4$ & $\epsilon=16$ \\ \hline
  NA&   \textbf{76.6\%} &      15.4\% &       15.3\% &       61.5\% &        41.7\% \\
  LGD&  76.3\% &      \textbf{73.9\%} &       \textbf{65.7\%} &       \textbf{74.8\%} &        \textbf{72.2\%} \\
  \hline
\end{tabular}
\end{table}

\begin{figure*}
\centering
\includegraphics[width=0.8\linewidth]{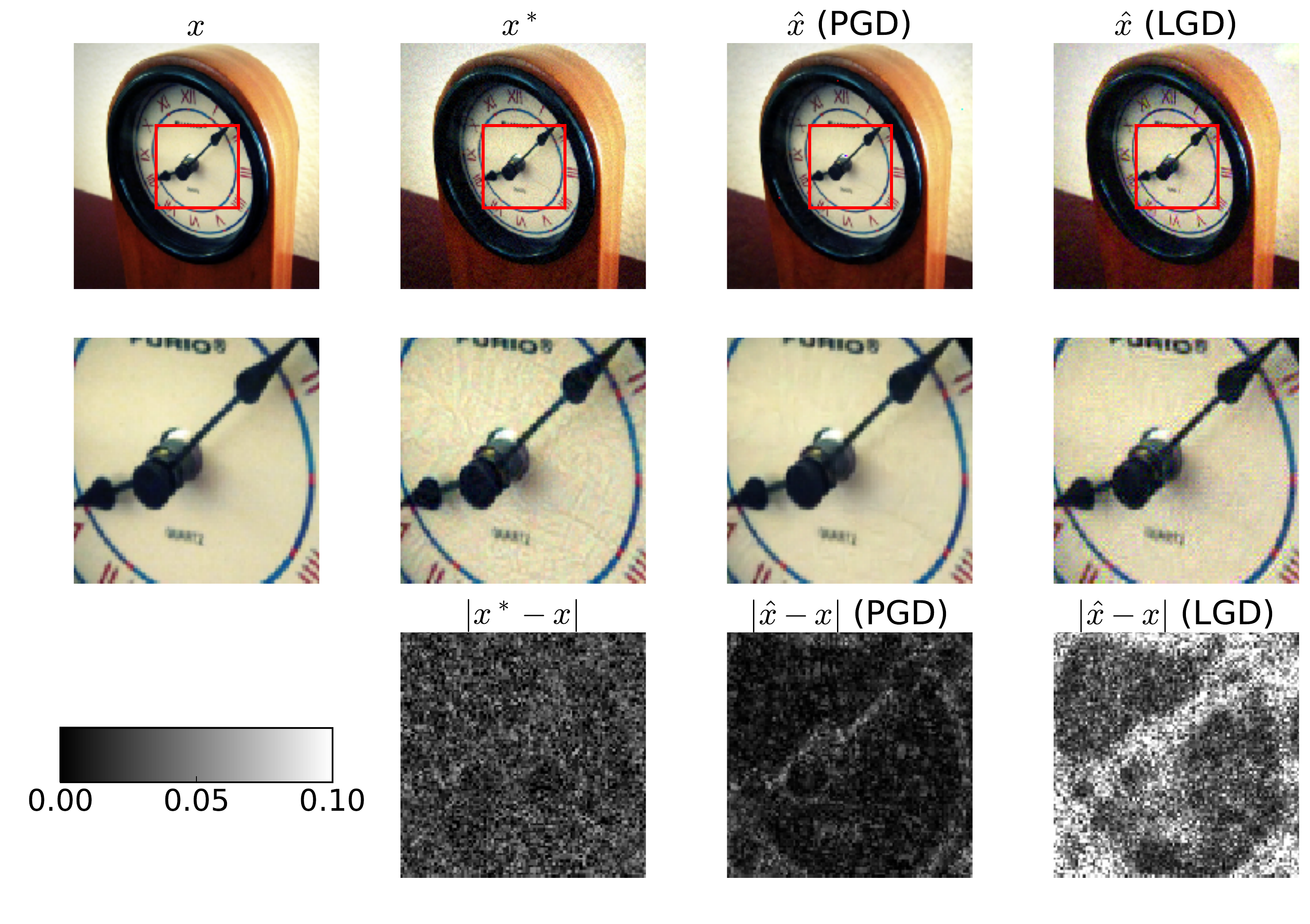}
\caption{$x$, $x^*$ and $\hat{x}$ denote the clean, adversarial and denoised images, respectively. The first row is an original image (1st column),  adversarial image (2nd column), denoised adversarial image generated by PGD (3rd column) and LGD (4th column). The second row shows zoomed in images. The third row visualizes the $L_1$ norm of differences between the original image and the last three images in the second row, respectively.}
\label{fig_demo}
\end{figure*}

\begin{figure}
\centering
\includegraphics[width=.8\linewidth]{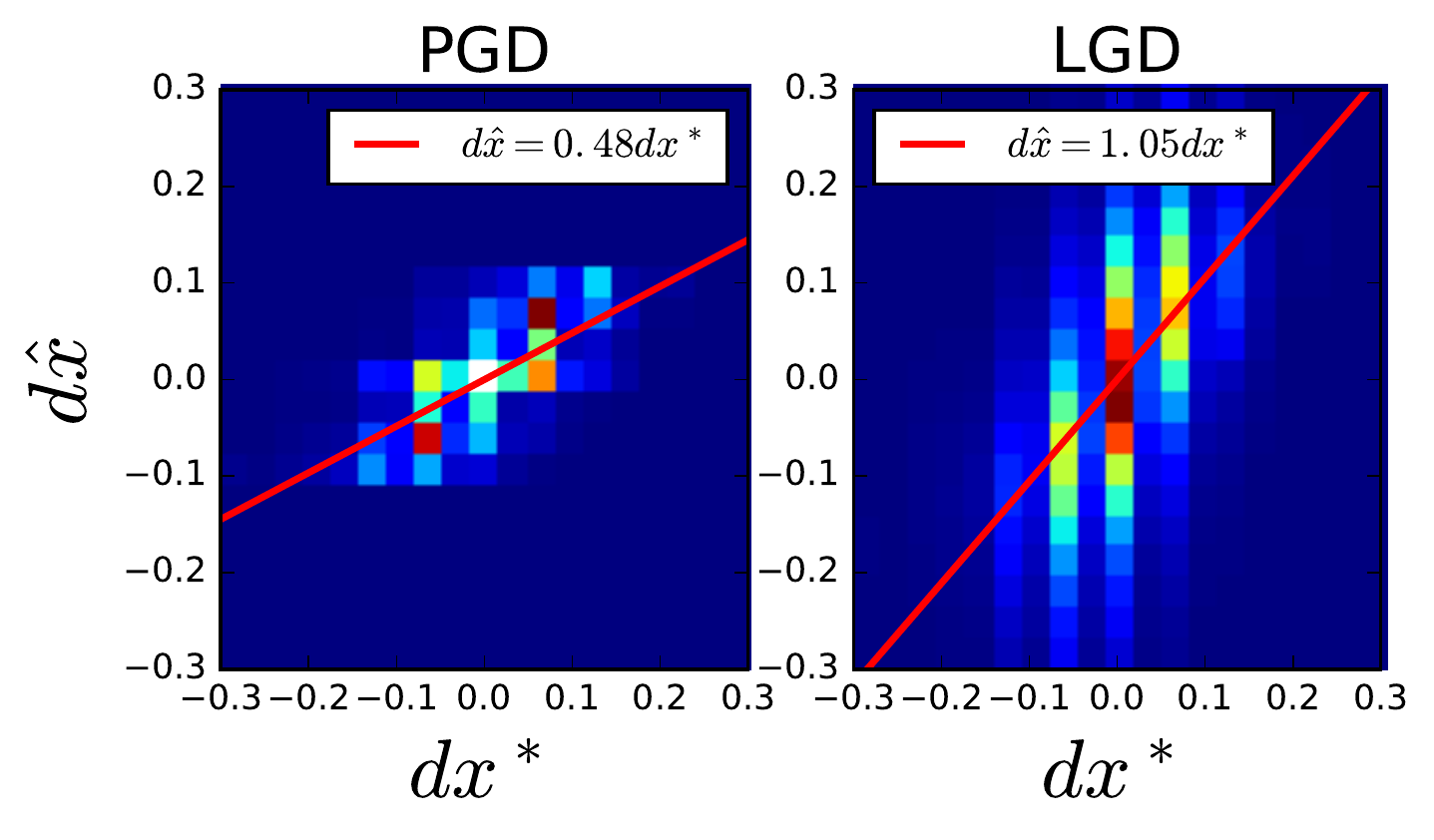}
\caption{The relationship between $dx^*$ and $d\hat{x}$ in PGD and HGD. }
\label{fig_over}
\end{figure}

\subsection{HGD as an anti-adversarial transformer}
\label{sec_entrap}

HGD is derived from a denoising motivation. However, HGD denoised images have larger pixel-level noise than adversarial images (see Figure \ref{fig_amp}), indicating that HGD even elevates the overall noise level.
This is also confirmed by the qualitative result in Figure \ref{fig_demo}. LGD does not suppress the total noise as PGD does, but adds more perturbations to the image.

To further investigate this issue, we plot the 2D histogram of the adversarial perturbation ($dx^* = x^* - x$) and the predicted perturbation ($d\hat{x}= x^*-\hat{x} $) by PGD and LGD (Figure \ref{fig_over}), where $x$, $x^*$ and $\hat{x}$ denote the clean, adversarial and denoised images, respectively. The ideal result should be $d\hat{x}=dx^*$, which means the adversarial perturbations are completely removed.

Two lines $d\hat{x} = kdx^*$ are fit for PGD and LGD, respectively (the red lines in Figure \ref{fig_over}). The slope of PGD's line is lower than 1, indicating that PGD only removes a portion of the adversarial noises.
In contrast, the slope of LGD's line is even larger than 1. Moreover, the estimation is very noisy, which leads to high pixel-level noise.

These observations suggest that HGD defends the target model by two mechanisms. First, HGD indeed reduces the adversarial noise level, which is revealed by the strong correlation between adversarial noise $dx^*$ and HGD induced perturbation $d\hat{x}$ (Figure \ref{fig_over}). Second, HGD adds to the image some “favorable perturbation” which defends the target model. In this sense, HGD can also be seen as an anti-adversarial transformer, which does not necessarily remove all the pixel-level noises but transforms the adversarial example to some easy-to-classify example.

\subsection{Results in NIPS adversarial defenses competition}
In NIPS 2017 competition track, Google Brain organized competition on adversarial attacks and defenses \footnote{\url{https://goo.gl/Uyz1PR}}.
The dataset used in this competition contains 5000 ImageNet-compatible clean images unknown to the teams. In the defenses competition, each team submits one solution, which are then evaluated on the attacks submitted by all teams. In total there are 91 non-targeted attacks and 65 targeted attacks. The evaluation is conducted on the cloud by organizers, and a normalized score is calculated based on the accuracy on all attacks. 

We used a FGD based solution. To train FGD, we gathered 14 kinds of attacks, all with $\epsilon=16$. Most of them were iterative attacks on an ensemble of many models (for details, please refer to supplementary file). We chose four pre-trained models (ensV3\cite{tramer2017ensemble}, ensIncResV2\cite{tramer2017ensemble}, Resnet152\cite{he2016deep}, ResNext101\cite{xie2016aggregated}) and trained a FGD for each one. The logits output of the four defended models were averaged, and the class with the highest score was chosen as the classification result. 

Our solution won the first place among 107 teams and significantly outperformed other methods (Table \ref{tab_comp}). Moreover, our model is much faster than the other top methods, measured by average evaluation time. 
\begin{table}[]
\centering
\caption{Results of the top five teams in NIPS defense competition. Time stands for average evaluate time.}
\label{tab_comp}
\begin{tabular}{@{}llll@{}}
\toprule
Team/Method &  Rank & Normalized Score & Time(s) \\ \midrule
iyswim   & 2 & 0.9235        & 121.83    \\
Anil Thomas & 3 & 0.9148        & 95.29     \\
erko   & 4 & 0.9120 & 86.44     \\
Stanford $\&$ Suns & 5 & 0.9106 & 127.39 \\
\hline
FGD (ours) & \textbf{1} & \textbf{0.9532}       & \textbf{50.24}      \\
\bottomrule
\end{tabular}
\end{table}


\section{Conclusion}
In this study, we discovered the error amplification effect of adversarial examples in neural networks and proposed to use the error in the top layers of the neural network as loss functions to guide the training of an image denoiser. This method turned to be very robust against both white-box and black-box attacks.
The proposed HGD has simple training procedure, good generalization, and high flexibility.

In future work, we aim to build an optimal set of training attacks. The denoising ability of HGD depends on the representability of the training set. In current experiments, we used FGSM and iterative attacks. Incorporating other different attacks, such as the attacks generated by adversarial transformation networks \cite{baluja2017adversarial}, probably improves the performance of HGD. It is also possible to explore an end-to-end training approach, in which the attacks are generated online by another neural network.

\vspace{-1ex}
\section*{Acknowledgements}
\vspace{-1ex}
\small{
The work is supported by the National NSF of China (Nos. 61620106010, 61621136008, 61332007, 61571261 and U1611461), Beijing Natural Science Foundation (No. L172037), Tsinghua Tiangong Institute for Intelligent Computing and the NVIDIA NVAIL Program, and partially funded by Microsoft Research Asia and Tsinghua-Intel Joint Research Institute.
}

\bibliography{adv}

\begin{thebibliography}{10}

\bibitem{baluja2017adversarial}
Shumeet Baluja and Ian Fischer.
\newblock Adversarial transformation networks: Learning to generate adversarial
  examples.
\newblock {\em arXiv preprint arXiv:1703.09387}, 2017.

\bibitem{biggio2013evasion}
Battista Biggio, Igino Corona, Davide Maiorca, Blaine Nelson, Nedim
  {\v{S}}rndi{\'c}, Pavel Laskov, Giorgio Giacinto, and Fabio Roli.
\newblock Evasion attacks against machine learning at test time.
\newblock In {\em Joint European Conference on Machine Learning and Knowledge
  Discovery in Databases}, pages 387--402, 2013.

\bibitem{carlini2017adversarial}
Nicholas Carlini and David Wagner.
\newblock Adversarial examples are not easily detected: Bypassing ten detection
  methods.
\newblock {\em arXiv preprint arXiv:1705.07263}, 2017.

\bibitem{das2017keeping}
Nilaksh Das, Madhuri Shanbhogue, Shang-Tse Chen, Fred Hohman, Li~Chen,
  Michael~E Kounavis, and Duen~Horng Chau.
\newblock Keeping the bad guys out: Protecting and vaccinating deep learning
  with jpeg compression.
\newblock {\em arXiv preprint arXiv:1705.02900}, 2017.

\bibitem{deng2009imagenet}
Jia Deng, Wei Dong, Richard Socher, Li-Jia Li, Kai Li, and Li~Fei-Fei.
\newblock Imagenet: A large-scale hierarchical image database.
\newblock In {\em IEEE Conference on Computer Vision and Pattern Recognition},
  pages 248--255, 2009.

\bibitem{dosovitskiy2016generating}
Alexey Dosovitskiy and Thomas Brox.
\newblock Generating images with perceptual similarity metrics based on deep
  networks.
\newblock In {\em Advances in Neural Information Processing Systems}, pages
  658--666, 2016.

\bibitem{goodfellow2014explaining}
Ian~J Goodfellow, Jonathon Shlens, and Christian Szegedy.
\newblock Explaining and harnessing adversarial examples.
\newblock {\em arXiv preprint arXiv:1412.6572}, 2014.

\bibitem{graese2016assessing}
Abigail Graese, Andras Rozsa, and Terrance~E Boult.
\newblock Assessing threat of adversarial examples on deep neural networks.
\newblock In {\em Machine Learning and Applications (ICMLA), 2016 15th IEEE
  International Conference on}, pages 69--74, 2016.

\bibitem{gu2014towards}
Shixiang Gu and Luca Rigazio.
\newblock Towards deep neural network architectures robust to adversarial
  examples.
\newblock {\em arXiv preprint arXiv:1412.5068}, 2014.

\bibitem{he2016deep}
Kaiming He, Xiangyu Zhang, Shaoqing Ren, and Jian Sun.
\newblock Deep residual learning for image recognition.
\newblock In {\em IEEE conference on computer vision and pattern recognition},
  pages 770--778, 2016.

\bibitem{he2016identity}
Kaiming He, Xiangyu Zhang, Shaoqing Ren, and Jian Sun.
\newblock Identity mappings in deep residual networks.
\newblock In {\em European Conference on Computer Vision}, pages 630--645,
  2016.

\bibitem{hinton2015distilling}
Geoffrey Hinton, Oriol Vinyals, and Jeff Dean.
\newblock Distilling the knowledge in a neural network.
\newblock {\em arXiv preprint arXiv:1503.02531}, 2015.

\bibitem{ioffe2015batch}
Sergey Ioffe and Christian Szegedy.
\newblock Batch normalization: Accelerating deep network training by reducing
  internal covariate shift.
\newblock In {\em International Conference on Machine Learning}, pages
  448--456, 2015.

\bibitem{johnson2016perceptual}
Justin Johnson, Alexandre Alahi, and Li~Fei-Fei.
\newblock Perceptual losses for real-time style transfer and super-resolution.
\newblock In {\em European Conference on Computer Vision}, pages 694--711.
  Springer, 2016.

\bibitem{kingma2014adam}
Diederik Kingma and Jimmy Ba.
\newblock Adam: A method for stochastic optimization.
\newblock {\em arXiv preprint arXiv:1412.6980}, 2014.

\bibitem{kurakin2016adversarial}
Alexey Kurakin, Ian Goodfellow, and Samy Bengio.
\newblock Adversarial machine learning at scale.
\newblock {\em arXiv preprint arXiv:1611.01236}, 2016.

\bibitem{lecun1998gradient}
Yann LeCun, L{\'e}on Bottou, Yoshua Bengio, and Patrick Haffner.
\newblock Gradient-based learning applied to document recognition.
\newblock {\em Proceedings of the IEEE}, 86(11):2278--2324, 1998.

\bibitem{meng2017magnet}
Dongyu Meng and Hao Chen.
\newblock Magnet: a two-pronged defense against adversarial examples.
\newblock In {\em Proceedings of the 2017 ACM SIGSAC Conference on Computer and
  Communications Security}, pages 135--147, 2017.

\bibitem{nayebi2017biologically}
Aran Nayebi and Surya Ganguli.
\newblock Biologically inspired protection of deep networks from adversarial
  attacks.
\newblock {\em arXiv preprint arXiv:1703.09202}, 2017.

\bibitem{osadchy2016no}
Margarita Osadchy, Julio Hernandez-Castro, Stuart Gibson, Orr Dunkelman, and
  Daniel P{\'e}rez-Cabo.
\newblock No bot expects the deepcaptcha! introducing immutable adversarial
  examples with applications to captcha.
\newblock {\em IACR Cryptology ePrint Archive}, 2016:336, 2016.

\bibitem{papernot2016transferability}
Nicolas Papernot, Patrick McDaniel, and Ian Goodfellow.
\newblock Transferability in machine learning: from phenomena to black-box
  attacks using adversarial samples.
\newblock {\em arXiv preprint arXiv:1605.07277}, 2016.

\bibitem{papernot2017practical}
Nicolas Papernot, Patrick McDaniel, Ian Goodfellow, Somesh Jha, Z~Berkay Celik,
  and Ananthram Swami.
\newblock Practical black-box attacks against machine learning.
\newblock In {\em ACM Asia Conference on Computer and Communications Security},
  pages 506--519, 2017.

\bibitem{papernot2016towards}
Nicolas Papernot, Patrick McDaniel, Arunesh Sinha, and Michael Wellman.
\newblock Towards the science of security and privacy in machine learning.
\newblock {\em arXiv preprint arXiv:1611.03814}, 2016.

\bibitem{papernot2016distillation}
Nicolas Papernot, Patrick McDaniel, Xi~Wu, Somesh Jha, and Ananthram Swami.
\newblock Distillation as a defense to adversarial perturbations against deep
  neural networks.
\newblock In {\em IEEE Symposium on Security and Privacy (SP)}, pages 582--597,
  2016.

\bibitem{rasmus2015semi}
Antti Rasmus, Mathias Berglund, Mikko Honkala, Harri Valpola, and Tapani Raiko.
\newblock Semi-supervised learning with ladder networks.
\newblock In {\em Advances in Neural Information Processing Systems}, pages
  3546--3554, 2015.

\bibitem{ridgeway2015learning}
Karl Ridgeway, Jake Snell, Brett Roads, Richard~S Zemel, and Michael~C Mozer.
\newblock Learning to generate images with perceptual similarity metrics.
\newblock {\em arXiv preprint arxiv:1511.06409}, 2015.

\bibitem{ronneberger2015u}
Olaf Ronneberger, Philipp Fischer, and Thomas Brox.
\newblock U-net: Convolutional networks for biomedical image segmentation.
\newblock In {\em International Conference on Medical Image Computing and
  Computer-Assisted Intervention}, pages 234--241, 2015.

\bibitem{szegedy2017inception}
Christian Szegedy, Sergey Ioffe, Vincent Vanhoucke, and Alexander~A Alemi.
\newblock Inception-v4, inception-resnet and the impact of residual connections
  on learning.
\newblock In {\em AAAI}, pages 4278--4284, 2017.

\bibitem{szegedy2016rethinking}
Christian Szegedy, Vincent Vanhoucke, Sergey Ioffe, Jon Shlens, and Zbigniew
  Wojna.
\newblock Rethinking the inception architecture for computer vision.
\newblock In {\em IEEE Conference on Computer Vision and Pattern Recognition},
  pages 2818--2826, 2016.

\bibitem{szegedy2013intriguing}
Christian Szegedy, Wojciech Zaremba, Ilya Sutskever, Joan Bruna, Dumitru Erhan,
  Ian Goodfellow, and Rob Fergus.
\newblock Intriguing properties of neural networks.
\newblock {\em arXiv preprint arXiv:1312.6199}, 2013.

\bibitem{tramer2017ensemble}
Florian Tram{\`e}r, Alexey Kurakin, Nicolas Papernot, Dan Boneh, and Patrick
  McDaniel.
\newblock Ensemble adversarial training: Attacks and defenses.
\newblock {\em arXiv preprint arXiv:1705.07204}, 2017.

\bibitem{vincent2008extracting}
Pascal Vincent, Hugo Larochelle, Yoshua Bengio, and Pierre-Antoine Manzagol.
\newblock Extracting and composing robust features with denoising autoencoders.
\newblock In {\em International Conference on Machine learning}, pages
  1096--1103, 2008.

\bibitem{xie2016aggregated}
Saining Xie, Ross Girshick, Piotr Doll{\'a}r, Zhuowen Tu, and Kaiming He.
\newblock Aggregated residual transformations for deep neural networks.
\newblock {\em arXiv preprint arXiv:1611.05431}, 2016.

\bibitem{zhang2017beyond}
Kai Zhang, Wangmeng Zuo, Yunjin Chen, Deyu Meng, and Lei Zhang.
\newblock Beyond a gaussian denoiser: Residual learning of deep cnn for image
  denoising.
\newblock {\em IEEE Transactions on Image Processing}, 2017.

\end{thebibliography}

\bibliographystyle{plain}
\end{document}